\documentclass[runningheads]{llncs}
\usepackage[T1]{fontenc}
\usepackage{siunitx}
\usepackage{graphicx,verbatim}
\usepackage{amsmath} 
\usepackage{booktabs}

\begin{document}
\title{Multimodal Shared Latent Representation of Narration, Microscope and iOCT Images for Phase Recognition in Vitreoretinal Surgery}
\titlerunning{Shared Latent Representation for Vitreoretinal Phase Recognition}

\author{Onur Izmitlioglu\inst{1}
\and Shervin Dehghani\inst{1, 2}
\and Tarek Ghannoum\inst{3}
\and Benedikt Schworm\inst{3}
\and Nassir Navab\inst{1}}
\authorrunning{Izmitlioglu et al.}

\institute{
Computer Aided Medical Procedures, Technical University of Munich, Germany \\
\email{\{onur.izmitlioglu, shervin.dehghani, nassir.navab\}@tum.de}
\and
SynthesEyes GmbH, Munich, Germany \\
\and
Department of Ophthalmology, LMU University Hospital, Munich, Germany \\
\email{t.ghannoum@med.uni-muenchen.de, benedikt.schworm@med.uni-muenchen.de}
}

% \author{**}
% \authorrunning{**}
% \institute{**}

\maketitle              % typeset the header of the contribution
\begin{abstract}
Surgical phase recognition is key to context-aware computer-assisted feedback in vitreoretinal procedures, yet the scarcity of synchronized multimodal intraoperative data, particularly microscope views and intraoperative OCT, limits approaches that aim to replicate the multimodal integration surgeons perform naturally. Surgical narration, by contrast, is abundantly available online and offers rich semantic supervision. Prior work has mainly explored pairwise contrastive learning (e.g., intraoperative OCT--microscope or microscope--narration), leaving the joint modeling of all three modalities largely unexplored. We introduce a framework that uses microscope views as a shared anchor to bridge surgical narrations and intraoperative OCT (iOCT) without requiring a fully synchronized tri-modal dataset, leveraging real microscope--narration videos and a synthetic dataset of synchronized microscope video and tool-aligned iOCT pairs. Contrastive alignment transfers structural priors from the synthetic domain to real videos lacking iOCT, and a dual-head \texttt{MS-TCN++} integrates the resulting embeddings for joint macro- and micro-phase prediction. Evaluated on real vitreoretinal surgeries, our framework improves macro-phase recognition over a zero-shot baseline (mean F1 $0.38\rightarrow0.53$) and provides an exploratory route to estimating fine-grained instrument--tissue measurements that are not directly observable in real microscope video alone; these micro-phase estimates are validated quantitatively on synthetic data and shown only qualitatively on real surgery. To our knowledge, this is the first work to unify microscope view, iOCT B-scans, and surgical narrations in a shared latent space for surgical phase recognition.
\end{abstract}

\section{Introduction}

Vitreoretinal surgery involves high-precision microsurgical procedures that demand exceptional skill and real-time decision-making. Enhancing a surgeon's contextual awareness through automated systems is critical for improving clinical outcomes, reducing cognitive load, and standardizing surgical training~\cite{Lalys2013,Quellec2014}. Surgical phase recognition serves as a foundational capability for such context-aware systems; however, it remains challenging in vitreoretinal procedures due to subtle tool-tissue interactions and the scarcity of annotated multimodal datasets.
We distinguish between two levels of surgical phases: macro-phases, corresponding to the conventional high-level stages of the procedure, and micro-phases, capturing fine-grained
instrument states such as forceps jaw closure and tool-to-retina distance.
Current AI-driven vitreoretinal surgery analysis often relies on single-modality video data, neglecting the depth-resolved structural insights provided by iOCT, which offers critical subsurface tissue information and tool-tissue proximity that video alone cannot capture.

\begin{figure}[t]
    \centering
    \includegraphics[width=0.9\textwidth]{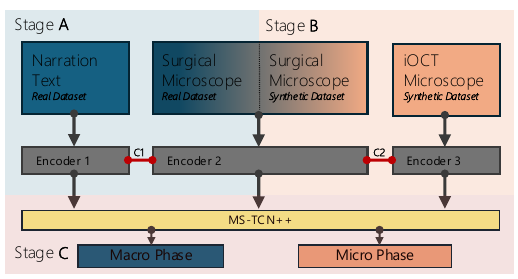}
    \caption{
        \textbf{Overview of the proposed framework.}
        We use three modality-specific encoders: \texttt{Encoder 1} (surgical narration), \texttt{Encoder 2} (microscope view), and \texttt{Encoder 3} (iOCT B-scans).
        \textbf{Stage A:} \texttt{Encoder 2} is aligned to the frozen \texttt{Encoder 1} via constraint \texttt{C1}, bringing microscope frames and narrations into a shared domain.
        \textbf{Stage B:} with \texttt{Encoder 2} now frozen, \texttt{Encoder 3} is trained via constraint \texttt{C2} to project iOCT B-scans into the same domain.
        \textbf{Stage C:} the resulting embeddings train the dual-head \texttt{MS-TCN++} for macro- and micro-phase prediction.
        At inference, because semantically similar modalities occupy neighboring regions of the latent space, the \texttt{MS-TCN++} predicts both macro- and micro-phases even when only the microscope view is available.
    }
    \label{fig:overview}
\end{figure}

Surgical narrations offer rich semantic supervision \cite{Yuan2024a}, yet datasets jointly containing narrations, microscope views, and iOCT B-scans are largely absent from the literature. Curating such a tri-modal dataset is itself a significant challenge, requiring synchronized acquisition across heterogeneous imaging systems in a demanding intraoperative setting. Moreover, clinical iOCT B-scans are not constrained to anatomically informative regions, and current technology does not support tool-following acquisition, which leaves much of the captured cross-sectional information contextually decoupled from the surgical procedure. To address this, we leverage a synthetic dataset of tool-aligned iOCT B-scans, ensuring that the acquired cross-sections are anchored to the surgical site and thus carry procedure-relevant context. Utilizing microscope views as a common anchor, we then bridge two complementary datasets, one of microscope--narration pairs and one of microscope--iOCT pairs, aligning all three modalities in a shared latent space without requiring a fully annotated tri-modal dataset.

Once the shared latent space is learned, supervised by surgical narrations and fine-grained annotations of the synthetic dataset, any single modality carries enough semantic structure to support inference about the others at inference time, as the joint supervision brings their representations into close proximity. For instance, a microscope view alone could in principle support an approximation of tool-to-retina distance without iOCT, to the extent that the two representations come to lie near each other in the latent space; we regard such cross-modal inference as an exploratory direction and evaluate it directly only where ground truth is available (Sec.~\ref{sec:results}).

% Dr. Simon Chen
\begin{figure}[tb]
    \centering
    \includegraphics[width=0.92\textwidth]{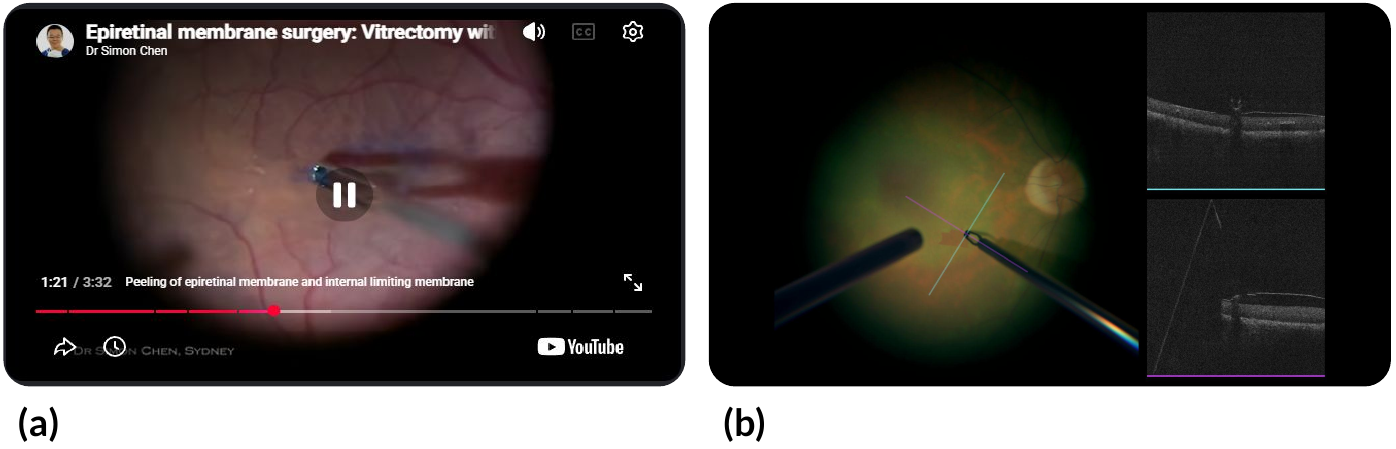}
    \caption{
        \textbf{Overview of the dataset.}
        \textbf{(a)} A sample frame from a youtube video taken from Dr. Simon Chen channel, containing the narration with the microscope view. \textbf{(b)} A sample frame from the synthetic dataset, containing the microscope view, the tool-aligned iOCT cross-sections, and forceps properties (tool-to-retina distance and forceps jaw closure).
    }
    \label{fig:dataset}
\end{figure}

\section{Related Works}
\subsubsection{Deep Learning in Surgical Contexts}
Deep learning, particularly CNNs and RNNs, shifted surgical workflow analysis toward robust, data-driven methods. \cite{Twinanda2016} introduced \textit{EndoNet} for laparoscopic videos, proving that deep models could outperform handcrafted features by learning directly from raw data. To better capture temporal dependencies, \cite{Czempiel2020} developed \textit{TECNO}, utilizing multi-stage convolutional networks, while \cite{Shinozuka2022} advanced clinical integration with real-time phase recognition for cholecystectomies. \cite{Demir2023} synthesized these trends, highlighting the shift toward multimodal, context-aware systems while noting that a lack of labeled data remains a significant bottleneck.

\subsubsection{Ophthalmic Surgery and Domain-specific Challenges}
While progress has been made in general surgery, ophthalmic procedures—particularly vitreoretinal—remain underrepresented. \cite{AlHajj2019} addressed this through the CATARACTS challenge for tool annotation, yet most research still prioritizes laparoscopic or robotic surgery, overlooking the unique visual traits of eye procedures. Recently, \cite{Hu2024} introduced OphNet, a large-scale benchmark for ophthalmic workflows, enabling more targeted, fine-grained temporal analysis. Despite this, the field still lacks narrated or context-rich data, a gap the current study aims to bridge. More recently, synthetic datasets have opened new avenues for multimodal learning in surgical settings.~\cite{rohrmoser2026toward} propose training on tool-aligned iOCT cross-sections paired with fundus images to estimate tool-to-retina distance.

\subsubsection{Language as Context in Surgical Modeling}
The integration of language into surgical video analysis has introduced vital semantic context. \cite{GuzmanGarcia2021} showed that speech-based phase recognition provides non-intrusive insights for skill assessment. Furthering the \textit{language as context} approach, \cite{Yuan2025} leveraged multimodal learning from video lectures, using narration as a rich supervisory signal for temporal alignment. Similarly,~\cite{gastager2026watch} reinforced that expert language is a crucial supervisory signal for enhancing video-based surgical understanding, training a video-language model with contrastive alignment to learn short-term spatio-temporal and multimodal representations for surgical phase detection, which are then passed to an \texttt{MS-TCN++} to capture long-range temporal relationships across the full video.

\subsubsection{Vision-Language Pretraining and Contrastive Learning}
Vision-language pretraining and contrastive learning have shifted surgical modeling toward zero-shot and few-shot capabilities. \cite{Yuan2024a} and \cite{Yuan2024b} introduced hierarchical models that align video frames with text to improve contextual transfer. This study specifically builds on \texttt{PeskaVLP}, which uses narrated content and contrastive alignment to bridge the visual-semantic gap and overcome annotation scarcity. Furthermore, frameworks like \texttt{IMAGEBIND}~\cite{Girdhar2023} demonstrate that contrastive objectives can align diverse modalities beyond text, such as audio and images, in a shared latent space, which underscores the versatility of modality-agnostic pretraining, supporting our objective of aligning OCT and microscope view representations for vitreoretinal surgery.

\section{Methodology}
\subsection{Macro- and Micro-Phase Definitions} 
In this work, we focus on epiretinal membrane (ERM) peeling as a representative vitreoretinal procedure, chosen for its well-defined sequential structure and the availability of both real and synthetic data in our dataset. Our schema follows a two-tier hierarchy: macro-phases for procedural structure and micro-phases for instrument-tissue interaction.
\subsubsection*{Macro-phase}
We decompose ERM peeling into five macro-phases, defined by keystep-level timestamps and established in consultation with vitreoretinal surgical experts to ensure clinical validity:
\begin{enumerate}
    \item \textbf{Anterior Preparation:} Initial surgical setup, including trocar insertion 
    and port placement.
    \item \textbf{Posterior Preparation:} Establishment of intraocular visualization and 
    core vitrectomy.
    \item \textbf{Membrane Identification:} Localization of the target membrane, typically 
    aided by chromodye staining.
    \item \textbf{Membrane Peeling:} Active removal of the epiretinal membrane through 
    forceps manipulation.
    \item \textbf{Retina Stabilization:} Post-peeling assessment and restoration, 
    including procedures such as air-fluid exchange.
\end{enumerate}
\subsubsection*{Micro-phase} Micro-phases characterize instrument-tissue interaction through two continuous attributes available in the synthetic dataset: the \textbf{tool-to-retina distance} and \textbf{forceps jaw closure}.

\subsection{Dataset Description}

The real dataset comprises 33 ERM peeling procedures sourced from online platforms, featuring full surgical microscope views (avg. 3 min) with varying resolutions. Whisper \cite{radford2023robust} generated three annotation granularities: \textbf{Abstract-level}: video-level semantic summaries, \textbf{Keystep-level}: timestamped milestones for macro-phase segmentation and \textbf{Narration-level}: synchronous transcriptions of real-time surgeon commentary. Keystep timestamps used for macro-phase segmentation were reviewed and, where needed, corrected by a vitreoretinal surgeon, while narration-level transcripts were kept as produced by Whisper.
Videos are split at the procedure level into 28 training procedures and 5 held-out procedures, with all frames of a given procedure kept within a single split. The 5 held-out procedures are used both to evaluate the Stage~A zero-shot baseline (Table~\ref{tab:baseline_metrics}) and to report final macro-phase performance (Table~\ref{tab:mstcn_cropped_metrics}); no hyperparameter tuning is performed on them.

For the synthetic data, five ERM peeling procedures were provided by SynthesEyes GmbH \footnote{https://syntheseyes.com}, covering membrane interaction and instrument manipulation. We additionally use two synthetic subretinal-injection sequences performed with a straight cannula. These sequences share the retinal setting but differ in instrument and task; they provide a contrasting procedural context that sharpens the phase boundaries of ERM peeling and act as hard negatives during Stage~B alignment (Sec.~\ref{sec:stageB}). In all synthetic sequences the iOCT cross-sections are continuously aligned to the primary instrument, tracking its movement throughout the procedure. For all synthetic sequences, accompanying metadata provides per-frame \textbf{tool-to-retina distance}, a normalized value in the range $[0, 1]$ representing the vertical proximity of the instrument tip to the retinal surface. The ERM peeling sequences additionally provide \textbf{forceps jaw closure}, a continuous value in the range $[0, 1]$ where $0$ corresponds to fully open and $1$ to fully closed; the subretinal-injection sequences use a straight cannula and therefore carry no jaw-closure attribute. A sample from the dataset is shown in Fig.~\ref{fig:dataset}.

\subsection{Stage A: Vision-Language Model}
\begin{figure}[tb]
    \centering
    \makebox[\linewidth]{\includegraphics[width=1.08\textwidth]{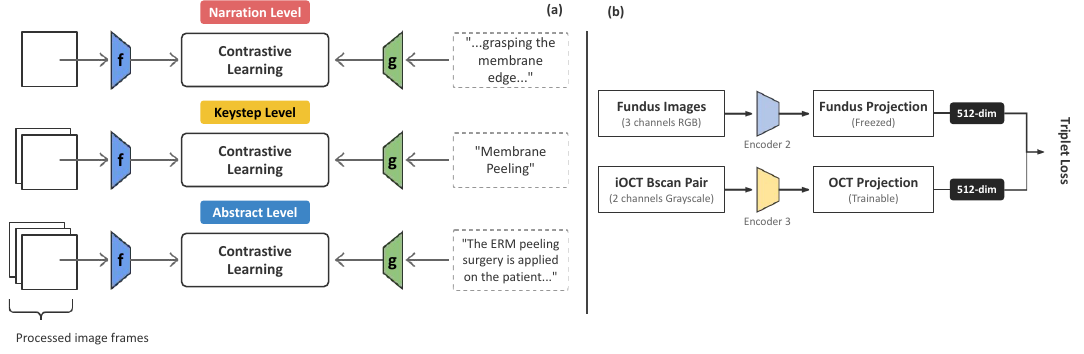}}
    \caption{
        \textbf{Overview of the models.}
        \textbf{(a)} Stage A: vision-language pretraining with hierarchical (narration-, keystep-, and abstract-level) text-image pairs. $f = \texttt{Encoder 2}$, a trainable \texttt{ResNet-50} microscope-image encoder; $g = \texttt{Encoder 1}$, a frozen \texttt{Bio\_ClinicalBERT} text encoder.
        \textbf{(b)} Stage B: microscope-iOCT alignment. $f = \texttt{Encoder 2}$, frozen from Stage A; $g = \texttt{Encoder 3}$, a trainable iOCT encoder. Both branches project to a shared 512-dimensional space and are aligned with a hard triplet objective (Sec.~\ref{sec:stageB}).
        }
    \label{fig:models}
\end{figure}

The baseline model for surgical phase recognition utilizes Vision-Language Pre-training (VLP) to derive semantic representations from microscope frames. This approach adapts \texttt{PeskaVLP} \cite{Yuan2024a} to the vitreoretinal domain. \textbf{Dual-Encoder Architecture} projects visual and textual inputs into a shared embedding space, enabling the alignment of surgical frames with descriptions through contrastive learning, as shown in Fig.~\ref{fig:models}.

The \textbf{text encoder} (\texttt{Encoder 1} in Fig.~\ref{fig:overview}) utilizes \texttt{Bio\_ClinicalBERT} \cite{Alsentzer2019}, pretrained on the \texttt{MIMIC-III} dataset to provide deep understanding of surgical terminology. It processes tokens through 12 transformer layers, summing the final four to capture both low-level and high-level semantics. During fine-tuning, the \texttt{BERT} parameters are frozen to maintain clinical knowledge, while the image encoder is trained to adapt to the vitreoretinal domain. The \textbf{image encoder} (\texttt{Encoder 2} in Fig.~\ref{fig:overview}) uses a \texttt{ResNet-50} backbone \cite{He2016}, initialized with \texttt{ImageNet} weights. Microscope frames are resized to $336 \times 336$ pixels and normalized. The architecture extracts spatial features through residual blocks, followed by global average pooling and a linear projection layer that maps the output to the 512-dimensional latent space.

\subsection{Stage B: Multimodal Alignment with iOCT Data}\label{sec:stageB}
To close the gap between synthetic iOCT data and real-world microscope videos, we adopt a dual-encoder architecture grounded in metric learning and cross-modal alignment principles~\cite{Schroff2015facenet,Kazerouni2025multimodal}, projecting both modalities into a unified 512-dimensional embedding space. \texttt{Encoder 2} remains frozen from Stage A to preserve its pretrained semantic procedural knowledge, while a trainable iOCT encoder (\texttt{Encoder 3} in Fig.~\ref{fig:overview}) is jointly optimized on synthetic scans, enriching the microscope representations with depth-resolved structural information and bridging the two modalities without requiring real iOCT data. The iOCT encoder is a CNN that maps each pair of tool-aligned cross-sections to this shared embedding space, with alignment learned via a \textbf{hard triplet loss} that pulls distances toward $0$ for temporally matched microscope-iOCT pairs and toward $1$ for unmatched pairs. Triplets are sampled by phase: given an anchor microscope frame and its temporally matched iOCT pair drawn from a synthetic sequence performed with one instrument (tool A), the negative is the iOCT pair of a randomly chosen frame from a sequence performed with the other instrument (tool B). Concretely, microscope frames showing forceps (ERM peeling) are contrasted against iOCT from cannula-based subretinal injection and vice versa, encouraging the encoder to represent instrument- and task-specific structure rather than generic retinal appearance.

\subsection{Stage C: Multi-Task Temporal Modeling}
Frame-level representations from Stage A and Stage B ignore temporal logic, leading to workflow inconsistencies and rapid phase oscillations \cite{Ding2024,ding2025mosformeraugmentingtemporalcontext} that median filtering cannot adequately resolve \cite{Demir2023}. While RNNs and LSTMs \cite{Lea2017} have traditionally modeled temporal sequences, they suffer from parallelization bottlenecks \cite{Lea2017,Ramesh2021}, vanishing gradients \cite{Bai2018}, and fixed-size memory constraints \cite{Ramesh2021}. TCNs \cite{Lea2017,Bai2018} overcome these by using dilated convolutions \cite{Yu2016} to exponentially expand the receptive field, capturing multi-scale procedural structures \cite{Yu2016,Czempiel2020}.

The \texttt{MS-TCN++} architecture \cite{Li2020} improves upon initial multi-stage refinement \cite{Farha2019} by employing dual dilated layers to capture both short-term landmarks and long-term constraints \cite{Czempiel2020,Farha2019,Li2020}. We implement a dual-head multi-task architecture to simultaneously address macro-phase classification and micro-phase regression.

\texttt{Head-1} regresses to five high-level phases using manual integer annotations from real surgical videos. While it successfully captures the overall procedural structure of ERM peeling, it lacks fine-grained details regarding instrument states or anatomical measurements.

\texttt{Head-2} utilizes synthetic videos and the accompanying metadata to regress to two continuous values, forceps jaw closure and tool-to-retina distance. Unlike the discrete classification of \texttt{Head-1}, this head characterizes subtle instrument-tissue interactions that are inherently unobservable in microscope footage but are observable in iOCT data.

\section{Experimental Results}\label{sec:results}

The implementation is done using \texttt{PyTorch 2.7.0} and \texttt{Python 3.11}. The codebase leverages GPUs with CUDA capabilities and standard deep learning libraries, including \texttt{NumPy 2.2.7} for numerical computations, \texttt{OpenCV 4.11.0} for video processing, and \texttt{Pillow 11.2.1} for image handling.

The held-out set of five surgical videos is used to assess the trained \texttt{PeskaVLP} model in order to create a baseline for surgical phase recognition. Phase predictions were made without any temporal modeling, following a zero-shot procedure: for each of the five macro-phases, a descriptive text prompt is designed and projected into the latent space via \texttt{Encoder 1}, while each microscope frame is independently projected into the same latent space via \texttt{Encoder 2}. The predicted phase for a given frame is then assigned as the class whose text embedding is closest to the frame embedding, with closeness measured via cosine similarity. The baseline model's performance is displayed in Table~\ref{tab:baseline_metrics}.

\begin{table}[htbp]
  \centering
  \small
  % --- LEFT TABLE ---
  \begin{minipage}[t]{0.48\textwidth}
    \centering
    \begin{tabular}{@{}lccc@{}}
      \toprule
      \textbf{Phase} & \textbf{Prec.} & \textbf{Rec.} & \textbf{F1} \\
      \midrule
      Anterior Prep.   & 0.59 & 0.38 & 0.46 \\
      Posterior Prep.  & 0.20 & 0.03 & 0.06 \\
      Membrane Id.     & 0.28 & 0.39 & 0.33 \\
      Membrane Peeling & 0.77 & 0.64 & 0.70 \\
      Retina Stab.     & 0.24 & 0.55 & 0.33 \\
      \midrule
      \textbf{Mean}    & \textbf{0.42} & \textbf{0.40} & \textbf{0.38} \\
      \bottomrule
    \end{tabular}
    \vspace{6pt}
    \caption{Baseline zero-shot macro-phase recognition performance after \textbf{Stage A}}
    \label{tab:baseline_metrics}
  \end{minipage}
  \hfill
  % --- RIGHT TABLE ---
  \begin{minipage}[t]{0.48\textwidth}
    \centering
    \begin{tabular}{@{}lccc@{}}
      \toprule
      \textbf{Phase} & \textbf{Prec.} & \textbf{Rec.} & \textbf{F1} \\
      \midrule
      Anterior Prep.   & 0.53 & 0.70 & 0.61 \\
      Posterior Prep.  & 0.50 & 0.84 & 0.63 \\
      Membrane Id.     & 0.31 & 0.32 & 0.31 \\
      Membrane Peeling & 0.76 & 0.68 & 0.72 \\
      Retina Stab.     & 0.42 & 0.36 & 0.38 \\
      \midrule
      \textbf{Mean}    & \textbf{0.50} & \textbf{0.58} & \textbf{0.53} \\
      \bottomrule
    \end{tabular}
    \vspace{6pt}
    \caption{Macro-phase recognition performance on real surgical videos with refined labels followed by \textbf{Stage C}}
    \label{tab:mstcn_cropped_metrics}
  \end{minipage}
\end{table}

For macro-phase recognition, the dual-head \texttt{MS-TCN++} model with label refinement outperformed the zero-shot baseline, achieving a mean F1 score of \textbf{53\%} with mean precision of \textbf{50.3\%} and mean recall of \textbf{57.9\%}. The performance breakdown by surgical phases is shown in Table~\ref{tab:mstcn_cropped_metrics}. The model shows particularly strong gains for Anterior and Posterior Preparation, while Membrane Identification and Retina Stabilization remain comparatively harder to recognize. This configuration jointly introduces temporal modeling, label refinement, and multi-task learning relative to the Stage~A baseline; disentangling their individual contributions requires dedicated ablations, which we leave to future work.

Micro-phase prediction was assessed quantitatively on the held-out synthetic ERM peeling sequences, where per-frame ground truth is available, and only qualitatively on real surgical videos. On synthetic data, the model achieved an MAE of \textbf{0.050 $\pm$ 0.001} for tool-to-retina distance and \textbf{0.167 $\pm$ 0.135} for forceps jaw closure. This ordering is counterintuitive, since the lower MAE for distance would ordinarily indicate the easier target. The two signals, however, have very different statistics. Tool-to-retina distance changes slowly and stays within a narrow band for most of the procedure, so a near-constant, over-smoothed prediction already yields a small absolute error while failing to follow the instrument's true vertical motion, an expected consequence of the missing depth cues in 2D imagery. Forceps jaw closure instead spans the full $[0,1]$ range with frequent open/close transitions; the model recovers the timing and direction of these transitions but incurs larger errors and high per-sequence variance ($\pm$0.135) around them. In this sense forceps closure is predicted more \emph{informatively}, since the clinically meaningful state changes are captured, even though its MAE is higher. On real videos the same qualitative pattern holds: forceps-state estimates vary meaningfully over time, whereas distance estimates stay stable but conservative. As no real-domain ground truth is available, these real-video micro-phase results should be read as exploratory rather than as validated measurements. Overall, discrete state changes appear learnable under weak cross-modal supervision, while continuous depth estimation remains challenging without explicit 3D priors.

\subsubsection*{Limitations.} Several limitations should be noted. The study uses 33 real videos and seven synthetic procedures (five ERM peeling, two subretinal injection), with results reported on a single procedure-level split (28 training, 5 held-out real videos). The synthetic data is also limited in coverage: it does not naturally span all macro-phases, so only a few sequences are available per phase and some phases are not represented at all, which constrains the micro-phase supervision. The macro-phase improvement in Table~\ref{tab:mstcn_cropped_metrics} reflects the combined effect of temporal modeling, label refinement, and multi-task learning; our current experiments do not isolate the individual contribution of the narration/microscope/iOCT alignment through ablations. Micro-phase targets are validated only on synthetic data, and the structure of the learned shared latent space is not analyzed directly, either geometrically (e.g., via cross-modal retrieval or matched-versus-unmatched embedding distances) or topologically (e.g., via persistent homology of how the modalities and phases are connected in the embedding). Our comparison is against a zero-shot vision-language baseline; supervised video-only temporal models and standard domain-adaptation methods are not included. Ablations per modality and training stage, stronger supervised baselines, real-domain micro-phase annotation, multi-split/multi-seed reporting, and quantitative latent-space analysis are the focus of ongoing work.

\section{Conclusion}
Our results suggest that the surgical microscope image can serve as a common anchor to bridge otherwise disjoint datasets, allowing structural priors learned from synthetic iOCT to propagate into real surgical video without ever requiring synchronized tri-modal recordings. As synthetic data generation matures and the curation of real narrated surgical videos expands, both the fidelity of learned representations and the overall performance are expected to improve substantially. Rigorous evaluation on larger and more diverse datasets, including real surgical videos annotated with forceps properties to directly assess the method's adaptability, will be essential to disentangle the contribution of each modality, inform more principled training strategies, and determine under which clinical conditions each modality provides the greatest benefit. Loss functions that more explicitly encourage cross-modal attention between the modalities, rather than only pulling matched pairs together, would also be a promising direction. This is a natural next step toward building richer, more robust multimodal representations for surgical AI, a form of cross-modal reasoning that surgeons perform implicitly throughout a procedure.

\section{Acknowledgments}
This work was supported by the Bavarian Research Foundation (BFS) under Grant AZ-1569-22. We gratefully acknowledge Hessam Roodaki and Ghazal Ghazaei for their support and helpful discussions.
% **

\subsubsection*{Competing interests} S.D. and N.N. are shareholders of SynthesEyes GmbH, the company that provided the synthetic dataset used in this work. All other authors declare that they have no competing interests.

% \newpage
\bibliographystyle{splncs04}
\bibliography{bibliography}

\end{document}